\documentclass[12pt]{article}
\usepackage{fullpage}
\usepackage{graphicx}
\usepackage{float}
\usepackage{amsmath}
\usepackage{amssymb}
\usepackage{amsthm}
\usepackage{url}
\usepackage{threeparttable}
\usepackage[hmargin=1.0in,vmargin=1.0in]{geometry}
\usepackage{authblk}
\usepackage{array}
\usepackage{multirow}
\usepackage{booktabs}
\usepackage{xcolor}

\usepackage{color}

\usepackage[markup=underlined]{changes}
\usepackage{todonotes}

\definechangesauthor[name={},color=red]{r2r}
\setauthormarkup{}

\usepackage{fancyhdr}
\usepackage{latexsym}
\usepackage{verbatim}
\usepackage{multirow}
\usepackage{caption} 
\usepackage{natbib}
\usepackage{listings}
\usepackage{titlesec}
\titleformat*{\section}{\LARGE\bfseries}
\titleformat*{\subsection}{\Large\bfseries}
\titleformat*{\subsubsection}{\large\bfseries}

\usepackage{lineno}
\usepackage{setspace}

\begin{document}

\title{Can 3D point cloud data improve automated body condition score prediction in dairy cattle?}

\author[1]{Zhou Tang}
\author[1]{Jin Wang}
\author[1]{Angelo De Castro}
\author[1]{Yuxi Zhang}
\author[2]{Victoria Bastos Primo}
\author[2]{Ana Beatriz Montevecchio Bernardino}
\author[3]{Gota Morota}
\author[4]{Xu Wang}
\author[2]{Ricardo C Chebel}
\author[1*]{Haipeng Yu}
\affil[1]{Department of Animal Sciences, Institute of Food and Agricultural Sciences, University of Florida, Gainesville, FL, 32611 USA }
\affil[2]{Department of Large Animal Clinical Sciences, University of Florida, Gainesville, FL, 32611 USA}
\affil[3]{Laboratory of Biometry and Bioinformatics, Department of Agricultural and Environmental Biology, Graduate School of Agricultural and Life Sciences, The University of Tokyo, Bunkyo, Tokyo 113-8657, Japan}
\affil[4]{Department of Agricultural and Biological Engineering, Institute of Food and Agricultural Sciences, University of Florida, Gainesville, FL, 32611 USA}

\date{}

\maketitle

\noindent 
$^{*}$ Corresponding author: haipengyu@ufl.edu\\ 

\noindent
Running title: Point clouds for BCS prediction\\ 


\noindent
ORCID: 
0000-0003-0111-2568 (ZT), 
0009-0001-4389-4396 (JW), 
0000-0002-8712-6153 (AD), 
0009-0007-7519-3512 (YZ), 
0009-0005-8731-3935 (VP),
0000-0003-1255-9066 (AB),
0000-0002-3567-6911 (GM), 
0000-0002-7144-6865 (XW),
0000-0002-9700-8089 (RCC), and 
0000-0002-8923-9733 (HY)\\

\noindent
Email address: 
zhou.tang@ufl.edu (ZT), 
jin.wang@ufl.edu (JW), 
decastro.a@ufl.edu (AD), 
yuxizhang@ufl.edu (YZ),
bastosprimo.v@ufl.edu (VP), 
montevecchiobe.a@ufl.edu (AB), \\
gmorota@g.ecc.u-tokyo.ac.jp (GM), 
xuwang1@ufl.edu (XW), 
rcchebel@ufl.edu (RCC), and 
haipengyu@ufl.edu (HY) \\

\newpage
\doublespacing
\section*{Abstract}
Body condition score (BCS) is a widely used indicator of body energy status and is closely associated with metabolic status, reproductive performance, and health in dairy cattle; however, conventional visual scoring is subjective and labor-intensive. Computer vision approaches have been applied to BCS prediction, with depth images widely used because they capture geometric information independent of coat color and texture. More recently, three-dimensional point cloud data have attracted increasing interest due to their ability to represent richer geometric characteristics of animal morphology, but direct head-to-head comparisons with depth image–based approaches remain limited. In this study, we compared top-view depth image and point cloud data for BCS prediction under four settings: 1) unsegmented raw data, 2) segmented full-body data, 3) segmented hindquarter data, and 4) handcrafted feature data. Prediction models were evaluated using data from 1,020 dairy cows collected on a commercial farm, with cow-level cross-validation to prevent data leakage. Depth image–based models consistently achieved higher accuracy than point cloud–based models when unsegmented raw data and segmented full-body data were used, whereas comparable performance was observed when segmented hindquarter data were used. Both depth image and point cloud approaches showed reduced accuracy when handcrafted feature data were employed compared with the other settings. Overall, point cloud–based predictions were more sensitive to noise and model architecture than depth image–based predictions. Taken together, these results indicate that three-dimensional point clouds do not provide a consistent advantage over depth images for BCS prediction in dairy cattle under the evaluated conditions.

\newpage
\section*{Introduction}
Body condition score (BCS) is a widely used indicator of body fat accumulation, reflecting nutritional status, health, fertility, and milk yield in dairy cattle \citep{ROCHE20095769, chebel2018association,  pinedo2022association, casaro2024association}. Routine monitoring of BCS is therefore essential for timely farm management decisions and animal welfare support. However, traditional visual scoring performed by trained experts is subjective, labor-intensive, and often inconsistent across assessors \citep{QIAO2021106143}. With continued global population growth, increasing milk production per cow is required to meet rising nutritional demands \citep{BRITT20183722}, placing greater pressure on the need for efficient and scalable farm management practices. An automated digital BCS system that provides objective, repeatable, and scalable measurements therefore has strong potential to support these management goals.

Early automated BCS assessment using computer vision primarily relied on red, green, and blue (RGB) images, often extracting handcrafted contour-based features from the hindquarter region \citep{BEWLEY20083439, AZZARO20112126, BERCOVICH20138047}. These approaches typically require manual annotation of anatomical keypoints, which limits scalability. The emergence of deep learning enabled end-to-end solutions by allowing neural networks to automatically learn relevant patterns from images, thereby eliminating the need for explicit feature design \citep{he2016deep}. Various neural network architectures trained on RGB images have demonstrated improved BCS prediction accuracy compared with handcrafted features, including convolutional neural networks \citep{ani9070470, 9272284, ani13020194} and transformers \citep{xue2023cat}. Low-cost RGB sensor data have also been explored using lightweight neural network models \citep{SIACHOS20242499, LI2025350}. Nevertheless, RGB-based approaches remain sensitive to environmental illumination, background clutter, and camera viewpoint, which can limit robustness across farms. Moreover, the absence of explicit geometric information can bias predictions due to variations in coat color and texture.

Depth sensors provide complementary geometric information, including body shape, contour, and volume, offering improved consistency across animals with different coat colors \citep{yu2021pig, BI2023100352}. Handcrafted features derived from depth data have commonly relied on anatomical keypoints, such as relative height differences \citep{SPOLIANSKY20167714, s20133705} and distances between keypoints \citep{KUZUHARA2015186, 10379607}. Three-dimensional depth information has further enabled the extraction of features based on regression parameters from fitted two-dimensional contours \citep{SONG20194294} and three-dimensional surface models \citep{zhao2020automatic}. Convex hull areas \citep{zhao2020automatic} and volumetric measurements \citep{LIU202016, s20133705} extracted from regions of interest have also been explored. Depth sensors are substantially less sensitive to illumination changes and shadows. Converting depth data into grayscale images—either using raw depth values \citep{YUKUN201910140, WINKLER2024100482} or segmented body regions such as hindquarters \citep{FISCHER20154465, agronomy9020090}—enables the application of convolutional neural networks for BCS prediction while minimizing the influence of coat texture.

Several studies have investigated converting depth data into explicit three-dimensional representations, which can theoretically capture richer geometric characteristics of animal morphology. From three-dimensional point cloud reconstructions, geometric features, including distance, area-based, and volumetric measurements, extracted from lateral and dorsal views \citep{MARTINS2020104054} or from the rump region alone \citep{KOJIMA2022104816} have been used to predict BCS. Three-dimensional point clouds have also been converted into two-dimensional convex hull distance maps for convolutional neural network–based modeling \citep{ZHAO2023107588}. More recently, deep learning architectures designed specifically for point clouds, such as PointNet \citep{qi2017pointnetdeeplearningpoint} and DGCNN \citep{wang2019dynamicgraphcnnlearning}, have been applied to livestock data \citep{SHI2023107666}. Point clouds can additionally support mesh reconstruction to quantify shape differences among individual animals \citep{ZHANG2023108307}. Despite these potential advantages, the benefits of point cloud–based approaches relative to depth image data alone remain unclear.

Therefore, the current study investigated whether three-dimensional point cloud representations can enhance BCS prediction in dairy cattle. We systematically compared depth image–based and point cloud–based approaches under four settings: 1) unsegmented raw data, 2) segmented full-body data, 3) segmented hindquarter data, and 4) handcrafted feature data. All settings employed cow-level data partitioning, ensuring that evaluations were conducted using identical train–validation–test splits and preventing data leakage. This work provides both methodological insights and practical guidance for deploying automated digital BCS systems in commercial farm settings.

\newpage
\section*{Materials and Methods}
\subsection*{Animal image acquisition}
The data used in this study were collected from a commercial dairy farm in central Georgia, USA. Dairy cows were imaged in November 2024 and February 2025 using a depth-sensing image acquisition system equipped with a RealSense D455 camera (Intel Corp., Santa Clara, CA, USA). The camera was installed 2515 mm above the ground in a chute system at the exit of the milking parlor, enabling fully automated image acquisition as cows returned to the pen after milking. Each cow was independently scored by two trained evaluators using a five-point BCS scale with 0.25 point increments \citep{FERGUSON19942695}. Only cows with exact agreement between the two scorers were retained for modeling. In total, 1,020 dairy cows, comprising 10,228 depth CSV files and corresponding RGB images, were included in the study. These cows ranged in age from 3 to 10 years and had BCS values from 2.0 to 4.25. All animal procedures conducted for data collection were approved by the University of Florida Institutional Animal Care and Use Committee (IACUC protocol number 202200000534).

\subsection*{Experimental settings and data preprocessing}
\begin{figure}[H]
    \centering
    \scalebox{0.7}{
    \includegraphics[width=\linewidth]{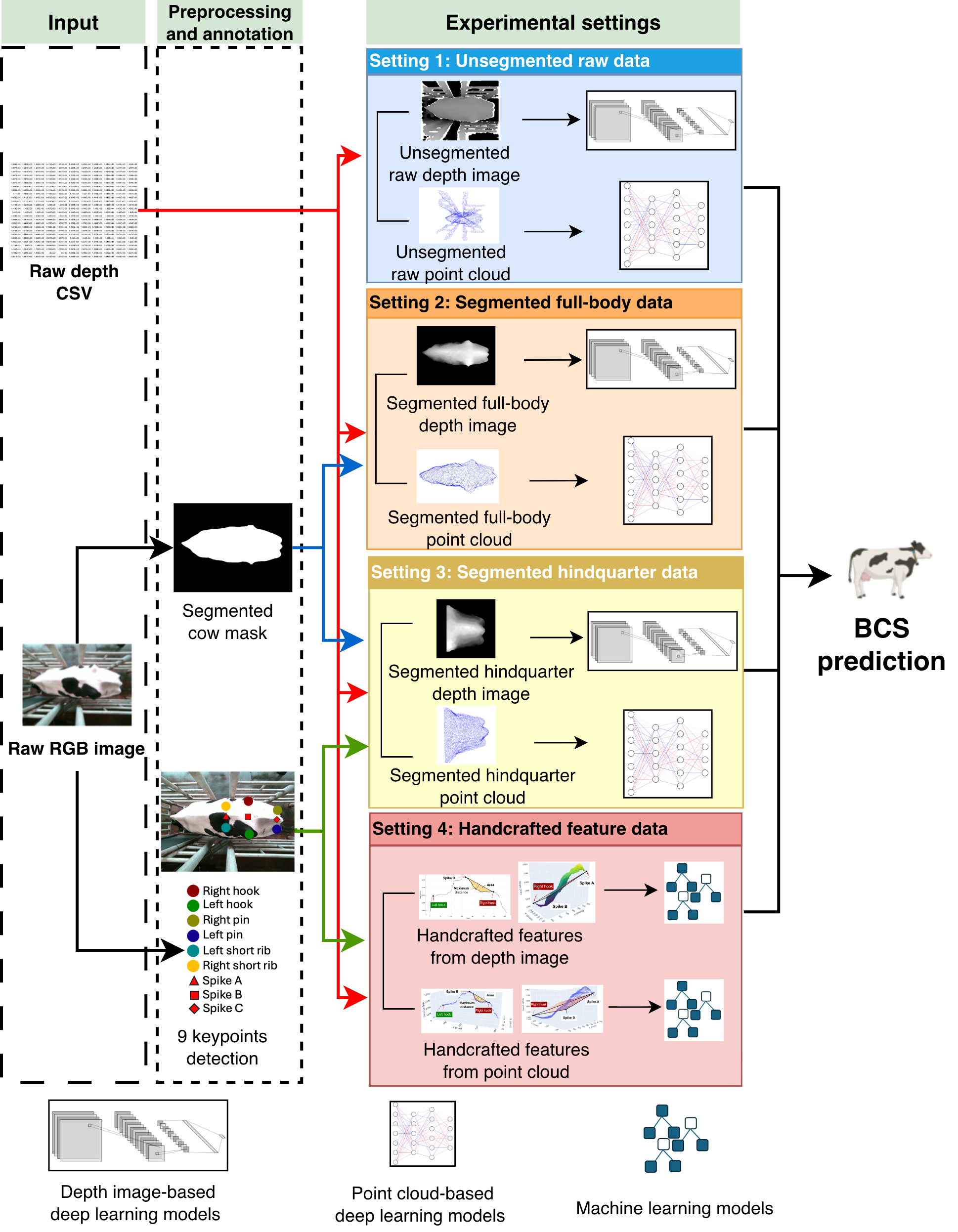} }
    \caption {Overview of the four experimental settings used in this study: 1) unsegmented raw depth data, 2) segmented full-body data, 3) segmented hindquarter data, and 4) handcrafted feature data. In setting~1, raw depth CSV files were converted into depth images and three-dimensional point clouds. RGB images were used for cow segmentation and keypoint detection, providing region-of-interest masks and anatomical landmarks for settings~2 and~3. The detected keypoints were further used in setting~4 to extract handcrafted geometric features, including distance and area features derived from two-point landmark pairs, as well as volumetric features derived from three-point landmark combinations. The depth image–based deep learning models included ResNet-18 and ConvNeXt, while the point cloud–based deep learning models included PointNet and DGCNN. The machine learning models applied to handcrafted features were random forest and LightGBM.}
    \label{fig:exp_overview}
\end{figure}

Four experimental settings were considered to systematically evaluate the impact of data representation on dairy cow BCS prediction (Figure~\ref{fig:exp_overview}): unsegmented raw data (setting 1), segmented full-body data (setting 2), segmented hindquarter data (setting 3), and handcrafted feature data (setting 4) derived from anatomical landmarks. Across all settings, depth images and point clouds were generated from the same raw depth measurements to ensure a fair and consistent comparison between representations.

\subsubsection*{Conversion of depth CSV to depth images and point clouds}
Raw depth measurements were recorded as two-dimensional CSV rasters. For depth image–based analysis, each depth raster was converted into a standardized height-map image using a fixed camera-to-ground distance of 2515 mm. Pixel-wise depth values were transformed into height above ground and normalized to an 8-bit grayscale range (0–255). Images were center-padded to preserve spatial alignment and standardized to uniform input dimensions. For point cloud–based analysis, the same depth CSV rasters were geometrically converted into three-dimensional point clouds using a pinhole camera model with calibrated intrinsic parameters and the same camera-to-ground reference \citep{Szeliski2022}. Each depth pixel was back-projected into camera coordinates to obtain its corresponding three-dimensional position. Only points with heights below the ground reference plane were retained to exclude invalid or background measurements. The resulting point clouds were exported as PLY files using the Open3D Python library \citep{zhou2018open3dmodernlibrary3d}. This unified conversion process ensured that depth images and point clouds were derived from identical raw measurements.

\subsubsection*{Image segmentation and keypoint detection}
To segment cows from the background across all 1,020 animals for downstream use in experimental settings 2 and 3, Grounded SAM was first used to generate initial cow masks without modeling training by applying the text prompt “cow” \citep{ren2024groundedsamassemblingopenworld} to a randomly selected subset of 130 cows, with one image per cow. These automatically resulting cow masks served as preliminary annotations and were manually reviewed and refined using LabelMe \citep{Russell2008LabelMe} to correct boundaries, remove spurious regions, and ensure consistent anatomical coverage. The finalized annotations were then used to train a lightweight YOLOv11n-seg model \citep{jocher_2025_15271204}. The 130 annotated images were randomly partitioned into 110 images for training, 10 for validation, and 10 for testing to develop and evaluate the model. Segmentation performance was assessed using precision, recall, the F1 score, and mean average precision at an intersection-over-union threshold of 0.5 (mAP@0.5). Precision and recall quantify the correctness and completeness of detected cow regions, respectively, while the F1 score represents their harmonic mean. The mAP@0.5 summarizes detection performance across confidence thresholds using an IoU threshold of 0.5, which is commonly adopted for object detection tasks in agricultural imaging. Once satisfactory segmentation performance was achieved, the trained YOLOv11n-seg model was used to automatically generate segmentation masks for all 1,020 cows across 10,228 depth images.

A keypoint detection model was developed to identify anatomical landmarks for experimental settings 3 and 4, which were manually annotated on RGB images using LabelMe. These include, left and right short rib points, left and right hook points, and left and right pin points. To predict these six anatomical keypoints across all 1,020 cows, a keypoint detection model based on YOLOv11s-pose was trained using a randomly selected subset of 877 cows, with one image per cow. These images were randomly partitioned into 70\% for training, 20\% for validation, and 10\% for testing to develop and evaluate the model. The YOLOv11s-pose model was trained with geometric and photometric augmentations implemented using the Albumentations library \citep{Buslaev2020Albumentations}. These augmentations included horizontal flipping (probability 0.5), affine transformations (probability 0.7; translation up to 2\% of image size, isotropic scaling within \(\pm 5\%\), and rotation within \(\pm 10^\circ\)), and appearance perturbations (probability of 0.2 each for brightness/contrast adjustment and additive Gaussian noise). Keypoint detection accuracy was evaluated using the percentage of correct keypoints (PCK) and root mean square error (RMSE). PCK measures the proportion of predicted keypoints whose Euclidean distance from the ground-truth location falls within a specified pixel threshold (k), providing a tolerance-based assessment of accuracy. RMSE quantifies the average localization error in pixels across all keypoints. Together, these metrics characterize both tolerance-dependent accuracy and absolute localization precision of anatomical landmark predictions. Once satisfactory keypoint detection performance was achieved, the trained model was used to automatically detect keypoints for all 1,020 cows across 10,228 RGB images.

\subsubsection*{Handcrafted geometric feature extraction}
A total of six predicted anatomical keypoints from the trained YOLOv11s-pose  model were refined to improve spatial accuracy in three-dimensional space (Figure~\ref{fig:exp_overview}). The hook and pin keypoints were adjusted by searching for the highest local point within a defined neighborhood around the initial predictions (30 pixels for hooks and 10 pixels for pins). Based on these refined landmarks, additional three anatomical keypoints were computed: spike A (the midpoint between the left and right short ribs), spike B (the midpoint between the adjusted left and right hooks), and spike C (the midpoint between the adjusted left and right pins) (Figure~S1), resulting in a total of nine anatomical keypoints. Three classes of handcrafted features were then extracted from both depth images and point clouds: maximum distance, area, and volume. 

\begin{figure}[H]
    \centering
    \includegraphics[width=\linewidth]{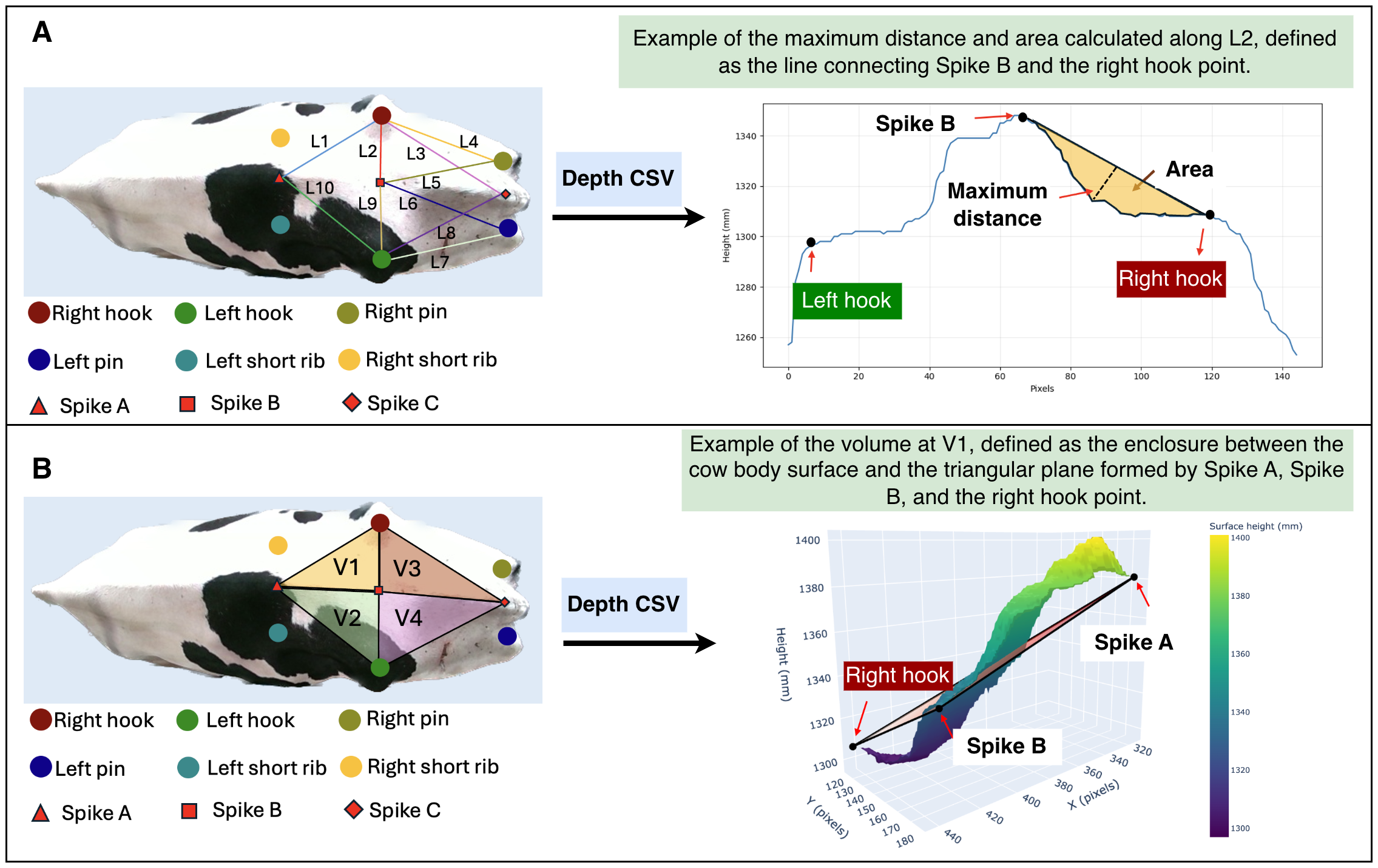}
    \caption{Illustration of handcrafted features extracted from depth images. A) The maximum distance and area were calculated from 10 lines (L1, L2, L3, L4, L5, L6, L7, L8, L9, and L10) derived by connecting the nine anatomical keypoints. Using L2 as an example, the maximum distance was defined as the height of the largest bulge of the body surface along L2, while the area was defined as the overall magnitude of body surface bulging along L2. B) The volume was separated into four volumes (V1, V2, V3, and V4). Using V1 as an example, it was defined as the volume between the cow body surface and the triangular anatomical plane defined by spike A, spike B, and the right hook point.}
    \label{fig:FeatureDepth}
\end{figure}

The maximum distance and area were calculated from 10 lines (L1, L2, L3, L4, L5, L6, L7, L8, L9, and L10) derived by connecting the nine anatomical keypoints (Figure~\ref{fig:FeatureDepth}). Using L2 as an illustration for depth images, the maximum distance was defined as the largest local protrusion of the cow’s body surface along L2. It was calculated as the greatest difference in height between the cow’s body surface and the straight line connecting spike B and the right hook point. In other words, this value captures how much the body surface bulges upward at its highest point along L2. The area represents the overall degree of body surface curvature along L2. It was calculated as the total area between the body surface and the straight line connecting spike B and the right hook point (Figure~\ref{fig:FeatureDepth}). Larger area values indicate a more pronounced and extended bulging of the body surface along this line, whereas smaller values indicate a flatter body surface. 

The volume was partitioned into four regions (V1, V2, V3, and V4) (Figure~\ref{fig:FeatureDepth}). Using V1 as an illustration for depth images, V1 represents the total volume enclosed between the cow’s body surface and the anatomical plane defined by spike A, spike B, and the right hook point (Figure~\ref{fig:FeatureDepth}). This volume was calculated by summing the vertical difference between the heights of the anatomical plane and the cow’s body surface over all points within the triangular region, with both heights measured relative to the ground \citep{ZHAO2023107588}. To estimate the height of all points within the triangular anatomical plane, the coordinates ($x$, $y$) and corresponding height ($z$) of spike A, spike B, and the right hook point were used to determine the plane equation $z = ax + by + c$ and obtain $a$, $b$, and $c$ using the NumPy package \citep{harris2020array}. These estimates were then used to compute the height of a given point ($x$, $y$) on the anatomical plane for volume calculation. Larger volume values indicate lower body mass distribution in this area.

\begin{figure}[H]
    \centering
    \includegraphics[width=\linewidth]{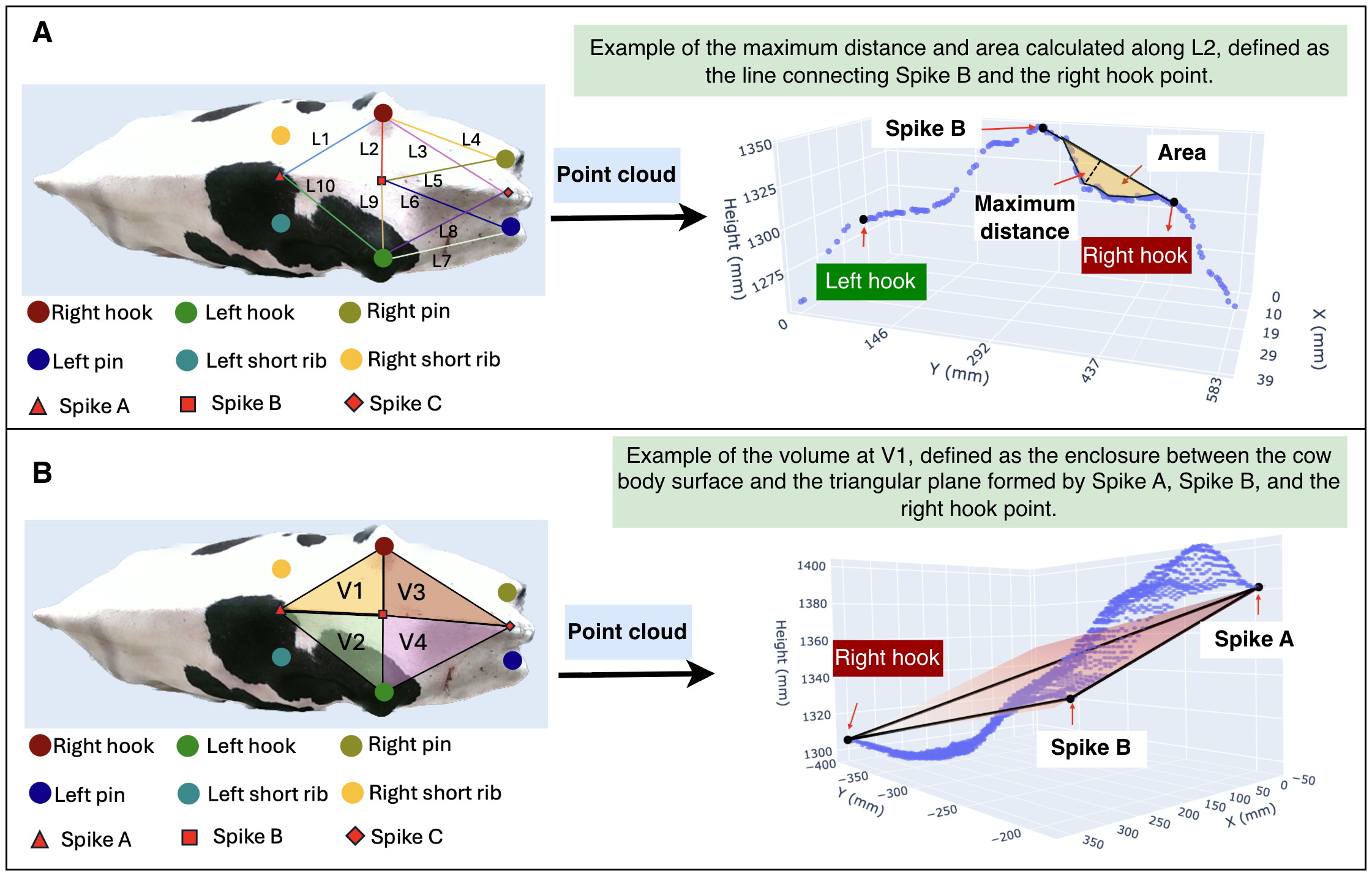}
    \caption{Illustration of handcrafted features extracted from point clouds. A) The maximum distance and area were calculated from 10 lines (L1, L2, L3, L4, L5, L6, L7, L8, L9, and L10) derived by connecting the nine anatomical keypoints. Using L2 as an example, the maximum distance was defined as the height of the largest bulge of the body surface along L2, while the area was defined as the overall magnitude of body surface bulging along L2. B) The volume was separated into four volumes (V1, V2, V3, and V4). Using V1 as an example, it was defined as the volume between the cow body surface and the triangular anatomical plane defined by spike A, spike B, and the right hook point.}
    \label{fig:FeaturePointClouds}
\end{figure}

For point cloud–based handcrafted features, depth rasters were first converted into calibrated three-dimensional point clouds. The same anatomical keypoints and 10 lines derived from the depth images were mapped into 3D space. Maximum distance, area, and volume \citep{ZHAO2023107588} features were then defined (Figure~\ref{fig:FeaturePointClouds}). This parallel feature extraction strategy ensured consistency between depth image and point cloud representations while capturing geometric characteristics in true three-dimensional space.

\subsection*{Body condition score prediction models}
Three types of predictive models were used to evaluate BCS prediction performance across different data representations: 1) deep learning models for depth images (ResNet-18 and ConvNeXt), 2) deep learning models for point clouds (PointNet and DGCNN), and 3) machine learning models for handcrafted features (random forest and LightGBM). All models were trained, validated, and tested within a unified experimental framework to ensure fair and reproducible comparisons. Together, these modeling strategies enabled a comprehensive evaluation of deep learning and machine learning approaches across image-based, point cloud–based, and handcrafted feature representations.

\subsubsection*{ResNet-18 and ConvNeXt}
Convolutional neural network–based classification models were implemented to predict BCS from depth images. Two representative architectures were selected: ResNet-18 \citep{he2016deep}, a widely used residual network serving as a classical convolutional neural network baseline, and ConvNeXt-Tiny \citep{liu2022convnext}, a modern convolutional architecture that incorporates design principles from vision transformers while retaining convolutional efficiency. Both networks were initialized with ImageNet-pretrained weights, and their final classification layers were replaced to match the number of BCS categories. Input depth images were resized and normalized using the ImageNet mean and standard deviation. Data augmentation, including random horizontal flipping and small-angle rotations, was applied during training to improve robustness to viewpoint variation. To mitigate class imbalance, a weighted random sampler was employed to oversample underrepresented BCS classes. Model parameters were optimized using the Adam optimizer with a ReduceLROnPlateau learning-rate scheduler. Label smoothing was applied to stabilize class predictions, and early stopping based on validation loss was used to prevent overfitting.

Hyperparameters were optimized using a Bayesian optimization search implemented in Optuna \citep{optuna2019}. The search space included the learning rate, weight decay, dropout rate, fully connected layer size, batch size, image resolution, number of training epochs, and early stopping patience (Table~\ref{predictionperform}). For each trial, the model was trained on the training set and evaluated on the validation set, with validation accuracy used as the optimization objective, while validation loss controlled early stopping and learning rate scheduler updates. Random seeds were fixed across Python, NumPy, and PyTorch to ensure deterministic and reproducible results. After hyperparameter optimization, the best performing configuration was retrained from scratch using the combined training and validation sets and the final model performance was evaluated on the test set. This workflow was applied identically to ResNet-18 and ConvNeXt-Tiny, enabling a consistent and fair comparison. Both ResNet-18 and ConvNeXt-Tiny were implemented using the Torchvision Python package \citep{paszke2019pytorch}. 

\subsubsection*{PointNet and DGCNN}
Deep learning models designed for unordered point sets were used to predict BCS from three-dimensional point cloud representations. Two architectures were evaluated: PointNet, which served as a baseline for direct point cloud learning \citep{qi2017pointnetdeeplearningpoint}, and DGCNN, which explicitly models local geometric relationships through dynamic graph construction \citep{wang2019dynamicgraphcnnlearning}. PointNet processes each point independently using shared multilayer perceptrons to extract local features, which are subsequently aggregated into a global shape descriptor via a symmetric max-pooling operation, enabling efficient capture of overall morphological characteristics. In contrast, DGCNN enhances feature learning by dynamically constructing k-nearest-neighbor graphs in feature space at each layer, thereby encoding both point-wise attributes and local neighborhood relationships that are sensitive to fine-scale geometric variations. This dynamic graph-based approach allows DGCNN to capture subtle surface curvature and contour differences associated with variation in body condition.

Each point cloud was optionally voxelized to homogenize point density, normalized to a unit sphere to remove scale and positional bias, and uniformly subsampled to a fixed number of points for model input. Data augmentation included random rotation around the vertical axis, isotropic scaling, and Gaussian jitter, which preserved anatomical structure while improving model robustness. Model hyperparameters, including learning rate, weight decay, dropout rate, voxel size, and the number of input points, were optimized using Bayesian optimization implemented in Optuna on the training and validation sets. The best performing configuration was retrained using the combined training and validation data and evaluated on the testing set. Random seeds were fixed throughout to ensure reproducibility. Both PointNet and DGCNN were implemented using the PyTorch framework \citep{paszke2019pytorch}.

\subsubsection*{Random forest and LightGBM}
Handcrafted features extracted from depth images and point clouds were modeled using machine learning approaches for tabular data. A total of 24 handcrafted features listed in Table S1 were included in prediction models.  A random forest classifier was implemented as a baseline ensemble model. Hyperparameters, including the number of trees, maximum tree depth, minimum leaf size, and feature sampling ratio, were optimized using the training and validation sets, with the configuration that maximized validation accuracy selected. The optimized model was then retrained on the combined training and validation data and evaluated on the test set. Feature importance scores derived from the trained ensemble were extracted to identify influential geometric predictors.

In the LightGBM model, a randomized hyperparameter search was conducted over the learning rate, number of leaves, maximum tree depth, subsampling ratios, regularization parameters, and class weights to address label imbalance. The best performing configuration was retrained on the combined training and validation data and evaluated on the test set. Feature importance rankings based on information gain were then extracted. The random forest and the LightGBM were implemented using the Python packages scikit-learn \citep{JMLRpedregosa11a} and lightgbm \citep{NIPS2017_6449f44a}, respectively. 

\subsection*{Model evaluation}
Model performance was evaluated using cow-level random subsampling cross-validation repeated five times to assess generalization to unseen animals and prevent data leakage \citep{WANG2025101483}. Final performance was reported as the mean and standard error of cow-level metrics across repeats. In each evaluation run, all images belonging to a given cow were assigned exclusively to either the training (70\%), validation (15\%), or testing (15\%) partition, ensuring that no cow appeared in more than one subset. This design eliminates identity leakage, which can otherwise artificially inflate performance when multiple images from the same animal are included across partitions. Within each training partition, a validation subset was created at the cow level and used for early stopping, learning rate scheduling, and hyperparameter tuning. Class distributions were monitored during partitioning, and cows were allocated to subsets to approximate stratification by BCS category while maintaining strict cow-level separation. All model hyperparameters were finalized prior to testing and were not adjusted using the test data.

To reflect real-world deployment scenarios in which multiple images may be collected per animal, performance was reported using cow-level rather than image-level metrics. For cow-level evaluation, predictions from all images belonging to the same cow were aggregated using majority voting to generate a single predicted BCS per animal, with each cow treated as an independent test instance. Cow-level accuracy was calculated under exact agreement (0 tolerance) as well as relaxed tolerance thresholds of 0.25 and 0.5 BCS units. In addition, two-sided t-tests were conducted to assess statistical significance between model performances. Overall, this evaluation framework provides a realistic and conservative assessment of predictive performance, closely reflecting operational conditions on commercial dairy farms where predictions are made for animals not previously observed during training.

\newpage
\section*{Results}
\subsection*{Overall dataset characteristics}
The distribution of BCS was consistent across the 2024 and 2025 datasets, comprising 475 and 545 observations, respectively (Figure~S2). Across both datasets, BCS values ranged from approximately 2.0 to 4.25, spanning cows from lean to well conditioned. The histograms showed unimodal distributions centered around BCS 3.0–3.5, which accounted for 601 of the 1,020 observations and are typical of well-managed dairy herds. Cows at the extremes of the scale, low BCS (2.0 or 2.25) and high BCS (4.0 or 4.25), represented 14.8\% of all observations. The 2025 distribution exhibited a slight rightward shift, consistent with its marginally higher mean BCS. Overall, the dataset was imbalanced toward moderate BCS levels, with relatively fewer observations at the extreme ends of the scoring range.

\subsection*{Image preprocessing results}
The segmentation model demonstrated consistently high detection performance across confidence thresholds. The F1 confidence curve exhibited a broad plateau, with F1 values exceeding 0.95 across a wide range of thresholds and reaching a maximum of 1.00 at a confidence threshold of approximately 0.93 (Figure S3). The corresponding precision–recall curve showed both precision and recall approaching 1.0, yielding an mAP@0.5 of 0.995. These results indicate that cow regions were reliably detected with minimal false positives and false negatives, providing a stable foundation for downstream feature extraction at scale while keeping manual effort manageable. 

The keypoint detection model achieved strong localization accuracy on the testing set. Quantitative evaluation using the PCK metric showed accuracies of 53.66\% at PCK@5, 88.01\% at PCK@10, 96.84\% at PCK@15, and 98.99\% at PCK@20. The overall RMSE across all landmarks was 6.97 pixels. Among the RMSE values for individual landmarks, the right hook showed an error of 4.65 pixels, the left hook 5.10 pixels, the right pin 5.29 pixels, and the left pin 5.64 pixels. In contrast, the left short rib at 8.82 pixels and the right short rib at 10.32 pixels exhibited larger errors. These higher errors are likely related to partial occlusion and lower visual contrast in these regions. Overall, the model achieved reliable landmark localization at moderate tolerance levels suitable for region of interest extraction and geometric feature computation.

\subsection*{Comparison in each setting}
\begin{figure}[H]
    \centering
    \includegraphics[width=\linewidth]{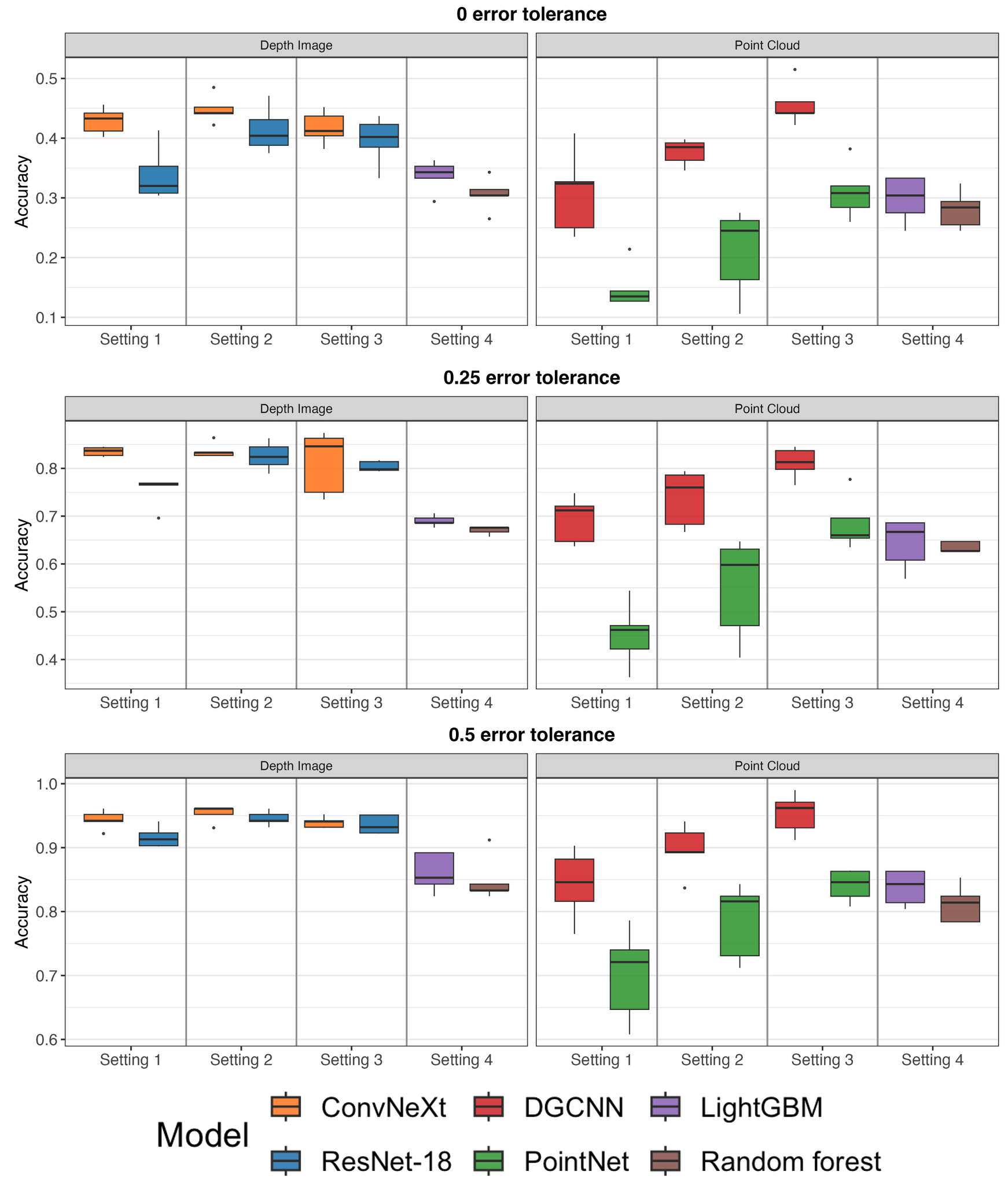}
    \caption{Boxplots of cattle-level prediction accuracy across four experimental settings under different error tolerance thresholds. Setting 1 corresponded to unsegmented raw data, setting 2 to segmented full-body data, setting 3 to segmented hindquarter data, and setting 4 to handcrafted feature data. The accuracy on the y-axis represents the proportion of cows that were correctly classified across different body condition score levels, averaged over five replicates. Box colors represent different model architectures. The left column corresponds to depth image inputs, and the right column corresponds to point cloud inputs.}
    \label{fig:accuracy_boxplot}
\end{figure}

Depth image–based models consistently achieved higher prediction accuracy than point cloud–based models in the unsegmented raw data (setting~1) and the segmented full-body data (setting~2) (Figure~\ref{fig:accuracy_boxplot}). Within setting~1, model choice had a substantial impact on performance within each data representation: ConvNeXt outperformed ResNet-18 for depth images, and DGCNN outperformed PointNet for point clouds. Across all tolerance levels, the depth image–based ConvNeXt model consistently outperformed the point cloud–based DGCNN model. For example, under a 0.25 BCS tolerance in setting~1, ConvNeXt achieved a mean accuracy of 0.84 compared with 0.69 for DGCNN, corresponding to an approximate 15\% relative improvement (Table~\ref{tab:overall_accuracy}). In setting~2, where full-body regions were segmented prior to modeling, the two depth image models achieved comparable performance, whereas differences between the two point cloud models remained evident. Depth image–based models continued to outperform point cloud–based models across all tolerance levels, with depth images yielding approximately 7\%, 10\%, and 5\% higher accuracy than point clouds under 0, 0.25, and 0.5 tolerance thresholds, respectively.

In contrast, no significant performance differences were observed between depth images and point clouds in the segmented hindquarter data (setting~3) or the handcrafted feature data (setting~4) (Figure~\ref{fig:accuracy_boxplot}). In setting~3, only differences between the two point cloud models were statistically significant. Although the depth image ConvNeXt model achieved the same accuracy (0.81) as the point cloud DGCNN model under the 0.25 tolerance, it showed slightly lower performance under the 0 tolerance (4\% lower) and the 0.5 tolerance (1\% lower). Two-sided t-tests comparing ConvNeXt and DGCNN in setting~3 yielded no statistically significant differences, with $p$-values of 0.05, 0.92, and 0.32 for the 0, 0.25, and 0.5 tolerance levels, respectively. In setting~4, no significant differences were observed between depth image–derived and point cloud–derived handcrafted features across models.

\subsection*{Comparison among different settings}
For depth image–based models, all depth image representations outperformed handcrafted feature–based models across all error tolerance levels (Figure~\ref{fig:accuracy_boxplot}). For example, LightGBM trained on handcrafted feature data (setting~4) achieved accuracies of 0.34, 0.69, and 0.86 under 0, 0.25, and 0.5 tolerance levels, respectively, whereas ConvNeXt trained on unsegmented raw depth data (setting~1) achieved accuracies of 0.43, 0.84, and 0.94 under the same conditions (Table~\ref{tab:overall_accuracy}). Comparisons among unsegmented raw depth data (setting~1), segmented full-body data (setting~2), and segmented hindquarter data (setting~3) revealed no statistically significant differences under relaxed tolerance levels (0.25 and 0.5). However, under the strict 0 tolerance, accuracy for the ConvNeXt-based segmented full-body data setting was significantly higher than that for the  unsegmented raw depth image ($p = 0.035$) and the segmented hindquarter data ($p = 0.020$). Overall, depth image–based approaches demonstrated robust and consistent performance across settings.

For point cloud–based models, point cloud representations did not consistently outperform handcrafted features across tolerance levels. For example, LightGBM trained on handcrafted features (setting~4) achieved accuracies of 0.30, 0.64, and 0.84 under 0, 0.25, and 0.5 tolerance, respectively, whereas DGCNN trained on unsegmented raw point clouds (setting~1) achieved accuracies of 0.31, 0.69, and 0.84, with no statistically significant differences observed between the two approaches. In DGCNN, segmented hindquarter data (setting~3) achieved the highest performance, with accuracies of 0.46, 0.81, and 0.95 under 0, 0.25, and 0.5 tolerance, respectively, followed by segmented cow point clouds (setting~2) and raw point clouds (setting~1). These results indicate that preprocessing strategies had a stronger influence on point cloud performance than model architecture alone.

\newpage
\section*{Discussion}
Image-based BCS assessment has been extensively studied in dairy cattle, with depth images consistently demonstrating advantages over RGB images due to their ability to capture body height and shape information while being less sensitive to illumination and coat color. More recently, three-dimensional point-cloud representations have attracted increasing interest for livestock phenotyping \citep{wang2026evaluating,morota2026artificial}. However, whether point clouds can provide measurable performance gains over depth images for BCS prediction has remained unclear. In this study, we systematically compared depth images and three-dimensional point clouds across four experimental settings using a large dataset comprising over 1,000 dairy cows. To minimize accuracy inflation, all models were evaluated using cow-level data partitioning, ensuring that images from the same animal never appeared in both the training and testing sets. Five repeated random subsampling cross-validation was conducted, with identical splits applied across all settings to enable fair and robust comparisons.

\subsection*{Comparison between depth images and point clouds}
Overall, the results demonstrate that both depth image–based and point cloud–based approaches can predict dairy cow BCS with high accuracy under relaxed tolerance thresholds (0.25 and 0.5 BCS units) when appropriate models are used, whereas performance under strict exact agreement (0 tolerance) remains more challenging. These findings are consistent with previous studies and, in many cases, match or exceed previously reported accuracies. For example, \citet{YUKUN201910140} reported accuracies of 0.45, 0.77, and 0.98 under 0, 0.25, and 0.5 tolerance using raw depth images. Our ConvNeXt-based unsegmented raw depth image model achieved comparable accuracy under 0 and 0.5 tolerance and higher accuracy under 0.25 tolerance (0.84). Similarly, for segmented full-body depth image setting, previous work using image-level data splits reported accuracies of 0.40, 0.81, and 0.97 under the same tolerance levels \citep{agronomy9020090}, whereas our ConvNeXt model achieved higher accuracy under 0 tolerance (0.45) and 0.25 tolerance (0.84), despite employing a stricter cow-level split. For point cloud–based models, our results obtained using a larger dataset and cow-level evaluation were comparable to previously reported performance \citep{ZHANG2023108307, SHI2023107666, ZHAO2023107588}. In contrast, handcrafted feature data yielded lower predictive accuracy than studies that used image-level data partitioning \citep{LIU202016, SONG20194294, zhao2020automatic}. This discrepancy is likely attributable to the elimination of identity leakage in our cow-based split, which has been shown to substantially reduce inflated accuracy estimates in livestock imaging studies \citep{WANG2025101483}.

Preprocessing strategies played a major role in shaping point cloud accuracy, but to a lesser extent for depth images. In the unsegmented raw data (setting~1) and segmented full-body data (setting~2), mean accuracies of depth image–based models were significantly higher than those of point cloud–based models across all three error tolerance levels, with lower variability. In contrast, in the segmented hindquarter data (setting~3) and the handcrafted feature data (setting~4), no significant differences were observed between depth images and point clouds. However, point cloud data must be subsampled to a fixed number of points before being input into deep learning models. When using the unsegmented raw data or segmented cow representations, a high density of points from background regions or BCS irrelevant body parts can occupy a substantial proportion of the subsampled points, thereby reducing the representation of informative anatomical features and degrading model performance. When focusing on the segmented hindquarter region, which contains the most physiologically relevant area for BCS prediction in dairy cattle, subsampling of point clouds is more likely to retain subtle and informative anatomical features. In contrast, handcrafted features derived from either depth images or point clouds compress complex dairy cow morphology into a small set of manually defined metrics, which resulted in a substantial performance gap for BCS prediction compared with deep learning–based representations.

Model architecture also exerted a strong influence on point cloud–based prediction accuracy. For example, DGCNN consistently outperformed PointNet across all experimental designs, reflecting the advantage of DGCNN’s edge-based convolutional operations in modeling local geometric neighborhoods compared with PointNet’s point-wise multilayer perceptron structure. In contrast, convolutional neural network–based depth image models exhibited greater robustness across preprocessing strategies. ConvNeXt generally outperformed ResNet-18, particularly when trained on raw depth images, although performance differences diminished under stronger segmentation. The lack of significant performance differences among depth image representations suggests that convolutional neural networks can effectively extract stable morphological cues from depth gradients even in the presence of background noise. Collectively, these results indicate that model architecture plays a critical role in depth-based BCS prediction, with higher-capacity architectures offering improved accuracy and reduced variability.

\subsection*{Limitations and future work}
Overall, this study found a general superiority of depth images over point clouds for BCS prediction, but several limitations warrant consideration. First, only top-view depth images were used for modeling. Incorporating rear-view or side-view perspectives may provide complementary anatomical information and further improve prediction accuracy \citep{BI2025104243}. Second, GPU memory constraints necessitated subsampling of point clouds prior to model training. Future work could explore more efficient subsampling strategies that preferentially retain informative regions, as well as GPU-efficient architectures capable of processing denser point clouds. In addition, standardizing point cloud representations across animals with varying postures remains an open challenge and may help reduce the risk of models learning identity-specific features rather than condition-related morphology.

In addition, depth image models benefited from transfer learning using ImageNet-pretrained weights, whereas point cloud models were trained from scratch. Developing large-scale pretraining strategies for point cloud representations may further enhance performance, analogous to the success of transfer learning in two-dimensional vision. From a practical perspective, depth image–based convolutional neural network models may offer greater reliability under typical farm conditions, where lighting variability, animal movement, and sensor placement are difficult to control. Point cloud–based models, while potentially more powerful for anatomically precise hindquarter analysis, may be better suited to controlled environments with high-quality segmentation and consistent point density. Hybrid approaches that integrate the stability of depth images with the geometric fidelity of point clouds represent a promising direction for future research.

\newpage
\section*{Conclusion}
This study systematically compared depth image and point cloud representations for automated BCS prediction using a large, cow-level dataset and a rigorous evaluation framework designed to prevent accuracy inflation. The results demonstrated that both depth image–based and point cloud–based models can achieve high predictive accuracy, comparable to or exceeding those reported in previous studies, particularly under relaxed error tolerance thresholds. Across experimental designs, depth images generally provided greater stability and lower performance variability than point clouds, whereas preprocessing strategies and model architecture exerted a stronger influence on prediction accuracy than data modality alone. Segmentation of physiologically relevant regions, such as the hindquarter, substantially improved point cloud performance and reduced the performance gap between representations. Collectively, these findings highlight that effective preprocessing and appropriate model selection are at least as important as the choice between depth images and point clouds for accurate BCS prediction. Overall, this work provides the first systematic and controlled comparison of depth image– and point cloud–based approaches for dairy cow BCS assessment. The results underscore the potential of geometry-aware learning frameworks for precision livestock phenotyping and suggest that combining the robustness of depth images with the geometric detail of point clouds may offer a balanced and effective solution. Future efforts will focus on multimodal fusion strategies, expanding dataset diversity across breeds and farm environments, and deploying these models in real-time, on-farm monitoring systems for automated BCS prediction.

\newpage
\section*{CRediT authorship contribution statement}

\textbf{Zhou Tang:} Investigation, Methodology, Software, Formal analysis, Visualization, Writing – original draft, Writing – review \& editing.
\textbf{Jin Wang}: Investigation, Methodology, Formal analysis, Writing – review \& editing.
\textbf{Angelo De Castro}: Investigation, Methodology, Writing – review \& editing.
\textbf{Yuxi Zhang}: Investigation, Writing – review \& editing.
\textbf{Victoria Bastos Primo}: Investigation, Writing – review \& editing.
\textbf{Ana Beatriz Montevecchio Bernardino}: Investigation, Writing – review \& editing.
\textbf{Gota Morota}: Methodology, Writing – review \& editing.
\textbf{Xu Wang}: Methodology, Writing – review \& editing.
\textbf{Ricardo C. Chebel}: Investigation, Methodology, Writing – review \& editing.
\textbf{Haipeng Yu}: Conceptualization, Investigation, Methodology, Writing – original draft, Writing – review \& editing, Supervision, Project administration, Funding acquisition.

\section*{Declaration of competing interest}
The authors declare no real or perceived conflicts of interest.

\section*{Funding statement}
This work was supported by the University of Florida startup funds to H.Y.



\newpage 
\bibliographystyle{apalike} 
\bibliography{bcs}

\newpage 
\section*{Tables}
\begin{table}[H]
\centering
\caption{Tuned hyperparameters and corresponding search ranges for depth image and point cloud models across different experimental settings.}
\label{predictionperform}
\scalebox{0.7}{
\begin{tabular}{l|l|l|l}
\hline\hline
\multicolumn{1}{c}{Model Input} &
\multicolumn{1}{c}{Model} &
\multicolumn{1}{c}{Hyperparameter} &
\multicolumn{1}{c}{Range} \\
\hline
\multirow{14}{*}{\textbf{Depth image}}
 & \multirow{7}{*}{ConvNeXt} 
 & Dense units number & (64, 128) \\
 & & Learning rate & $(1\times10^{-5}, 1\times10^{-3})$ \\
 & & Dropout probability & (0.1, 0.25) \\
 & & Weight decay & $(1\times10^{-4}, 5\times10^{-4})$ \\
 & & Label smoothing & (0.1, 0.25) \\
 & & Early stopping patience & (5, 10) \\
 & & Batch size & (32, 64) \\
\cline{2-4}
 & \multirow{7}{*}{ResNet-18}
 & Dense units number & (64, 256) \\
 & & Learning rate & $(1\times10^{-5}, 1\times10^{-3})$ \\
 & & Dropout probability & (0.1, 0.25) \\
 & & Weight decay & $(1\times10^{-4}, 5\times10^{-4})$ \\
 & & Label smoothing & (0.1, 0.25) \\
 & & Early stopping patience & (5, 10) \\
 & & Batch size & (32, 64) \\
\hline
\multirow{21}{*}{\textbf{Point cloud}}
 & \multirow{11}{*}{DGCNN}
 & Learning rate & $(1\times10^{-4}, 3\times10^{-3})$ \\
 & & Dropout probability & (0.1, 0.5) \\
 & & Weight decay & $(1\times10^{-5}, 1\times10^{-3})$ \\
 & & Label smoothing & (0.0, 0.2) \\
 & & Early stopping patience & (5, 10) \\
 & & Batch size & (8, 32) \\
 & & $k$ (kNN) & (10, 30) \\
 & & Embedding dimension & (256, 2048) \\
 & & Number of points & (1024, 20480) \\
 & & Voxel size & (1, 20) \\
 & & Subsample size & \{15000, 20000, 40000, 80000\} \\
\cline{2-4}
 & \multirow{10}{*}{PointNet}
 & Learning rate & $(1\times10^{-4}, 3\times10^{-3})$ \\
 & & Dropout probability & (0.1, 0.5) \\
 & & Weight decay & $(1\times10^{-5}, 1\times10^{-3})$ \\
 & & Label smoothing & (0.0, 0.2) \\
 & & Early stopping patience & (5, 10) \\
 & & Batch size & (8, 64) \\
 & & Embedding dimension & (256, 2048) \\
 & & Number of points & (1024, 20480) \\
 & & Voxel size & (1, 20) \\
 & & Subsample size & \{15000, 20000, 40000, 80000\} \\
\hline
\multirow{15}{*}{\textbf{Handcrafted feature}}
 & \multirow{6}{*}{Random Forest}
 & n\_estimators & \{200, 400, 800\} \\
 & & max\_depth & \{None, 8, 12, 16, 24\} \\
 & & min\_samples\_split & \{2, 5, 10\} \\
 & & min\_samples\_leaf & \{1, 2, 4\} \\
 & & max\_features & \{\texttt{sqrt}, \texttt{log2}, 0.5\} \\
 & & class\_weight & \{None, \texttt{balanced}\} \\
\cline{2-4}
 & \multirow{9}{*}{LightGBM}
 & n\_estimators & \{300, 600, 1000\} \\
 & & Learning rate & \{0.05, 0.1\} \\
 & & num\_leaves & \{31, 63, 127\} \\
 & & max\_depth & \{-1, 8, 12\} \\
 & & min\_child\_samples & \{10, 20, 50\} \\
 & & bagging\_fraction & \{0.7, 0.9, 1.0\} \\
 & & feature\_fraction & \{0.7, 0.9, 1.0\} \\
 & & lambda\_l2 & \{0.0, 1.0, 5.0\} \\
 & & class\_weight & \{None, \texttt{balanced}\} \\
\hline
\end{tabular}}
\end{table}

\newpage 
\begin{table}[H]
    \centering
    \caption{Body condition score prediction accuracy across three experimental settings. Results were based on repeated random subsampling cross-validation under three error tolerance levels (0, 0.25, and 0.5 body condition score units). SE denotes standard error.}
    \label{tab:overall_accuracy}
    \resizebox{\textwidth}{!}{
    \begin{tabular}{l|l|l|ll|ll|ll}
        \hline
        \hline
        \multirow{2}{*}{Experimental setting} &
        \multirow{2}{*}{Input} &
        \multirow{2}{*}{Model} &
        \multicolumn{2}{c}{Error 0} &
        \multicolumn{2}{c}{Error 0.25} &
        \multicolumn{2}{c}{Error 0.5} \\
        \cline{4-9}
        & & &
        \multicolumn{1}{c}{Mean} &
        \multicolumn{1}{c}{SE} &
        \multicolumn{1}{c}{Mean} &
        \multicolumn{1}{c}{SE} &
        \multicolumn{1}{c}{Mean} &
        \multicolumn{1}{c}{SE} \\
        \hline
        \multirow{4}{*}{\textbf{Setting 1: Unsegmented raw data}}
            & \multirow{2}{*}{Depth image}
                & ResNet-18 & 0.34 & 0.02 & 0.75 & 0.01 & 0.92 & 0.01 \\
            &   & ConvNeXt  & 0.43 & 0.01 & 0.84 & 0.00 & 0.94 & 0.01 \\ \cline{2-9}
            & \multirow{2}{*}{Point cloud}
                & PointNet  & 0.15 & 0.02 & 0.45 & 0.03 & 0.70 & 0.03 \\
            &   & DGCNN     & 0.31 & 0.03 & 0.69 & 0.02 & 0.84 & 0.02 \\
        \hline

        \multirow{4}{*}{\textbf{Setting 2: Segmented full-body data}}
            & \multirow{2}{*}{Depth image}
                & ResNet-18 & 0.41 & 0.02 & 0.83 & 0.01 & 0.95 & 0.00 \\
            &   & ConvNeXt  & 0.45 & 0.01 & 0.84 & 0.01 & 0.95 & 0.01 \\ \cline{2-9}
            & \multirow{2}{*}{Point cloud}
                & PointNet  & 0.21 & 0.03 & 0.55 & 0.05 & 0.79 & 0.03 \\
            &   & DGCNN     & 0.38 & 0.01 & 0.74 & 0.03 & 0.90 & 0.02 \\
        \hline

        \multirow{4}{*}{\textbf{Setting 3: Segmented hindquarter data}}
            & \multirow{2}{*}{Depth image}
                & ResNet-18 & 0.40 & 0.02 & 0.80 & 0.00 & 0.94 & 0.01 \\
            &   & ConvNeXt  & 0.42 & 0.01 & 0.81 & 0.03 & 0.94 & 0.00 \\ \cline{2-9}
            & \multirow{2}{*}{Point cloud}
                & PointNet  & 0.31 & 0.02 & 0.68 & 0.03 & 0.84 & 0.01 \\
            &   & DGCNN     & 0.46 & 0.02 & 0.81 & 0.01 & 0.95 & 0.01 \\
        \hline

        \multirow{4}{*}{\textbf{Setting 4: Handcrafted feature data}}
            & \multirow{2}{*}{Depth image}
                & Random Forest & 0.31 & 0.01 & 0.67 & 0.00 & 0.85 & 0.02 \\
            &   & LightGBM      & 0.34 & 0.01 & 0.69 & 0.01 & 0.86 & 0.01 \\ \cline{2-9}
            & \multirow{2}{*}{Point cloud}
                & Random Forest & 0.28 & 0.01 & 0.64 & 0.00 & 0.81 & 0.01 \\
            &   & LightGBM      & 0.30 & 0.02 & 0.64 & 0.02 & 0.84 & 0.01 \\
        \hline
    \end{tabular}}
\end{table}

\newpage
\section*{Figures}

\newpage 

\newpage 

\newpage 

\clearpage
\setcounter{figure}{0}
\renewcommand{\thefigure}{S\arabic{figure}}
\setcounter{table}{0}
\renewcommand{\thetable}{S\arabic{table}}
\section*{Supplemental Material}

\begin{figure}[H]
    \centering  
    \includegraphics[width=\linewidth]{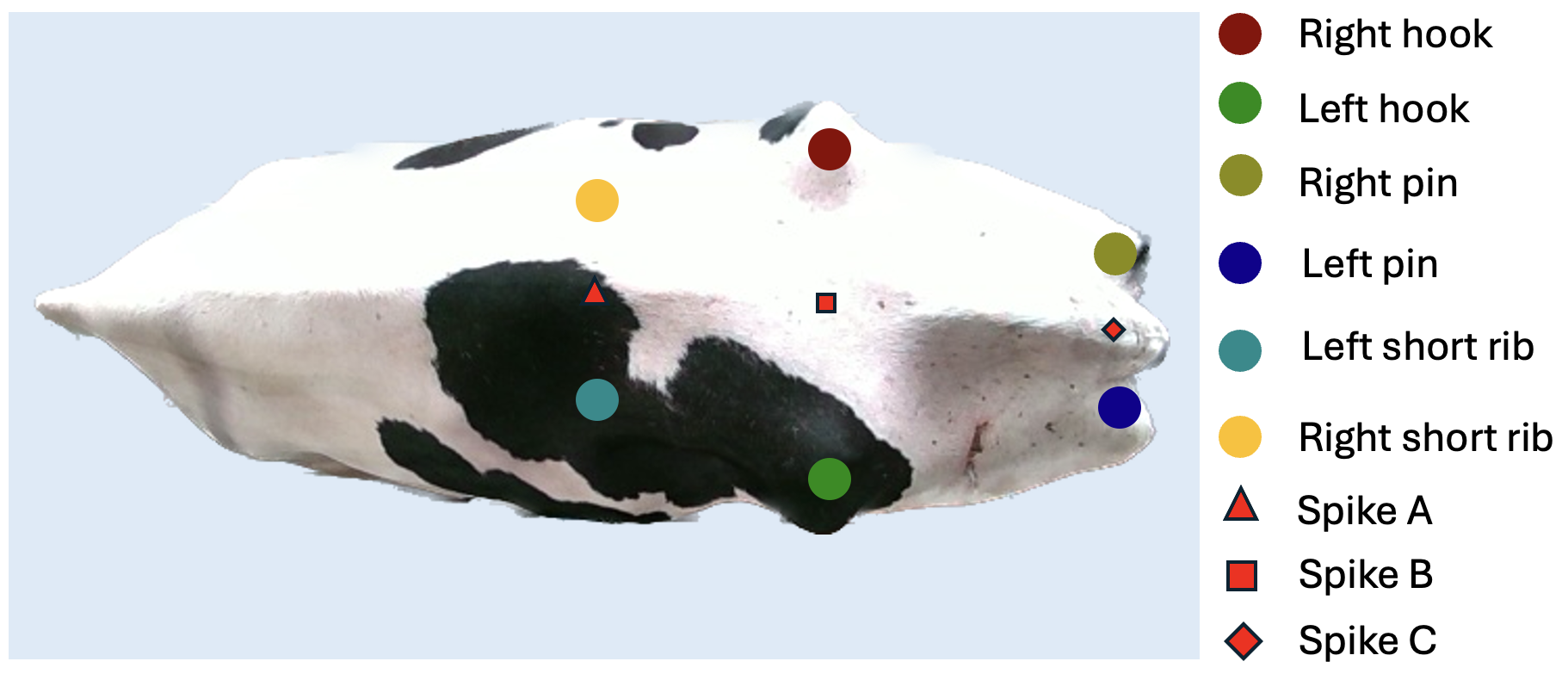}
    \caption{Keypoint detection and derived spike landmarks. The figure shows nine anatomical keypoints identified by the keypoint detection model. Derived spike landmarks were computed from these keypoints and used as reference points for downstream feature extraction and quantitative analysis.} 
    \label{Keypoint}
\end{figure}

\newpage 
\begin{figure}[H]
    \centering
    \includegraphics[width=\linewidth]{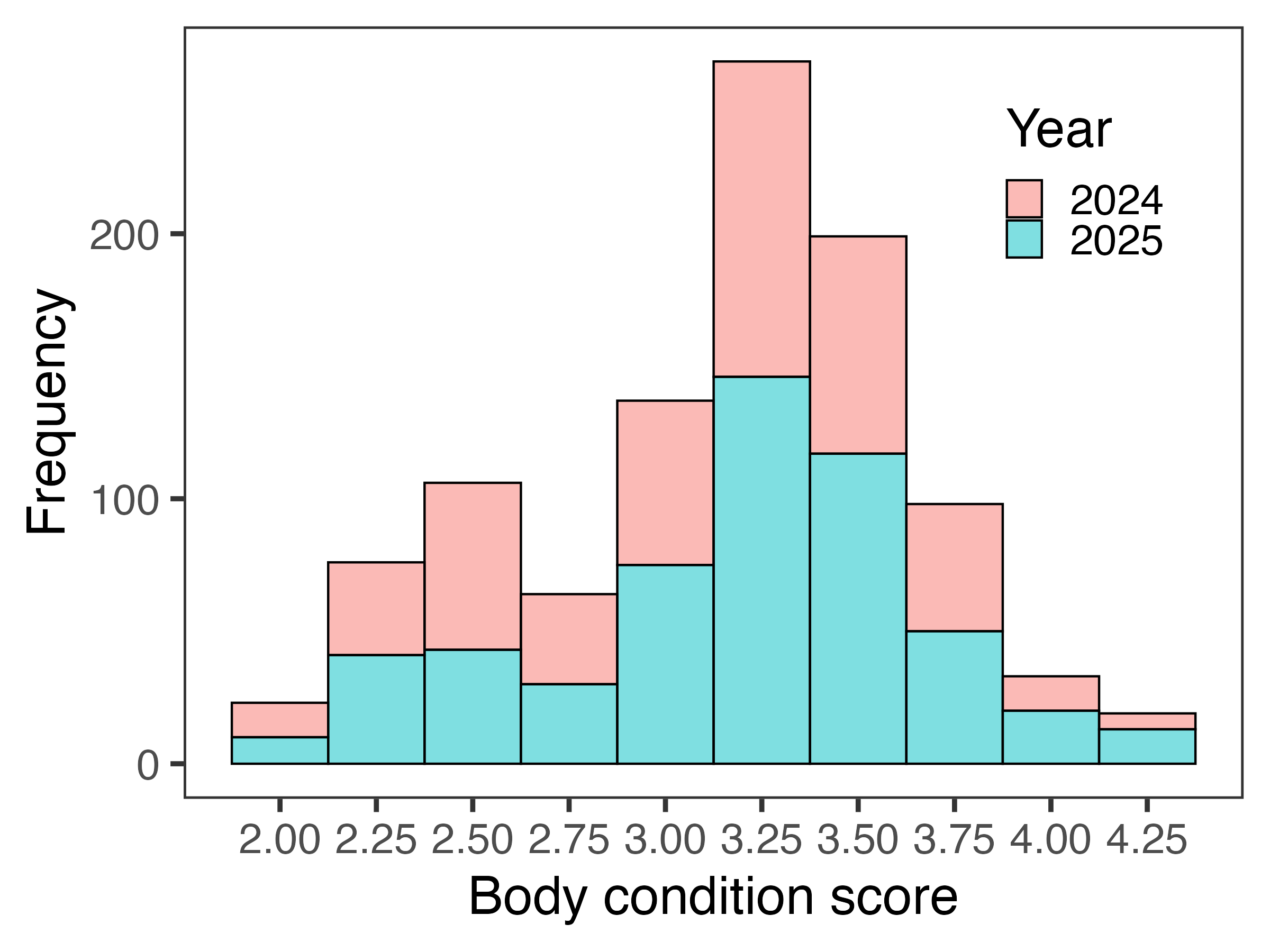}
    \caption{Distribution of body condition score data. Observations collected in 2024 and 2025 are shown in different colors in the histogram.}
    \label{fig:BCS_distribution}
\end{figure}

\newpage
\begin{figure}[H]
    \centering  
    \includegraphics[width=\linewidth]{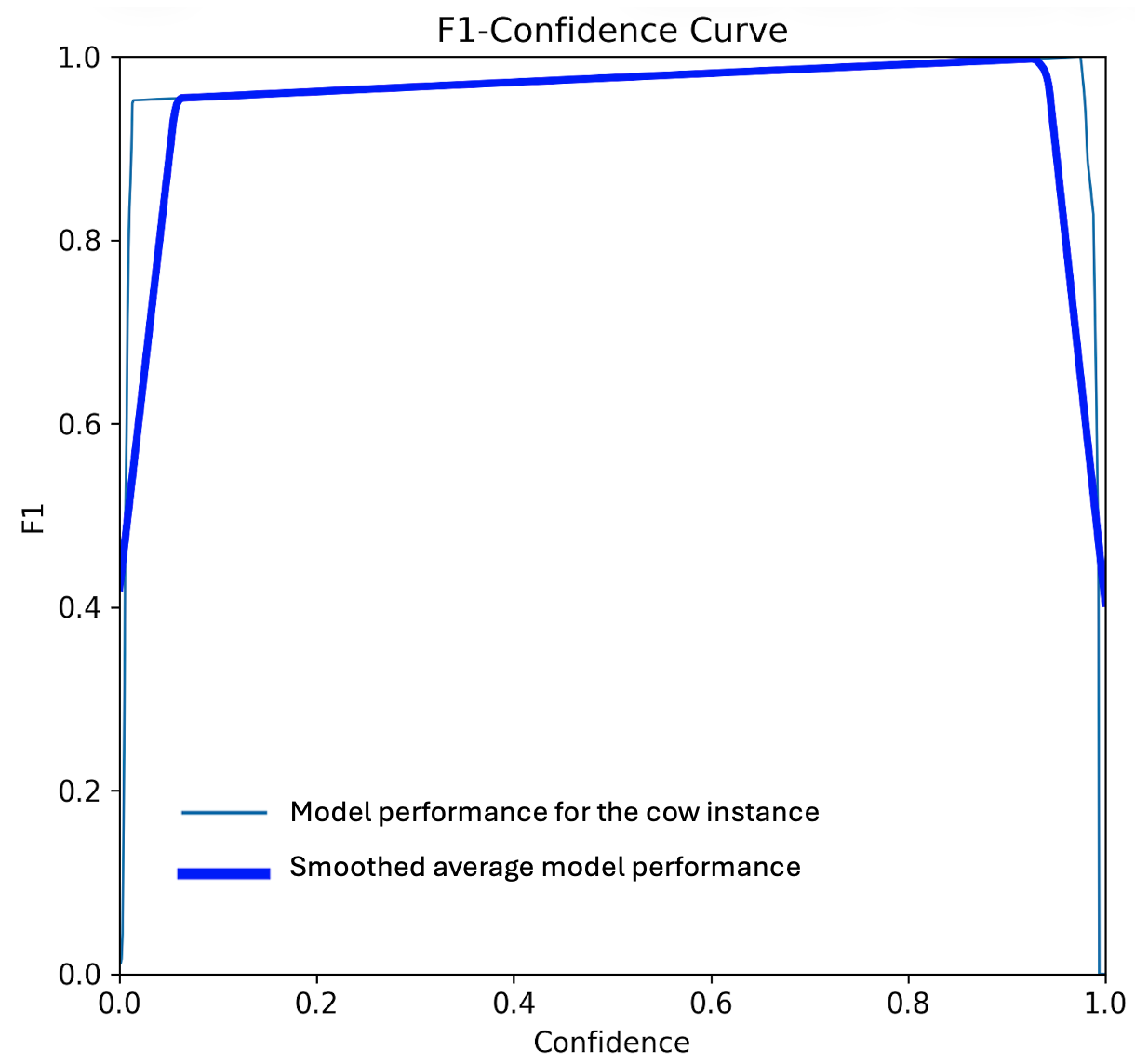}
    \caption{F1–confidence curve for the YOLO segmentation model on the test set. The F1 score (y-axis) was evaluated across different confidence thresholds (x-axis). The thin line represents the performance for the cow instance, while the thick line indicates the smoothed average performance across confidence thresholds. The maximum F1 score (1.00) is achieved at a confidence threshold of 0.925, which was selected as the optimal operating point for subsequent evaluations.} 
    \label{F1}
\end{figure}

\newpage 
\begin{table}[ht]
\centering
\caption{A list of handcrafted features derived from depth images and point cloud data that were used in prediction models in experimental setting 4. MaxDist refers to the maximum distance.}
\label{tab:handcrafted_features}
\scalebox{0.55}{
\begin{tabular}{l l l}
\hline
Feature type & Feature name & Description of handcrafted feature \\
\hline

\multirow{10}{*}{Maximum distance}
& MaxDist (L1)  & Maximum perpendicular distance between the cow body surface curve and line L1 connecting spike A and the right hook point  \\
& MaxDist (L2)  & Maximum perpendicular distance between the cow body surface curve and L2 connecting Spike B and the right hook point \\
& MaxDist (L3)  & Maximum perpendicular distance between the cow body surface curve and line L3 connecting Spike C and the right hook point \\
& MaxDist (L4)  & Maximum perpendicular distance between the cow body surface curve and line L4 connecting right pin and the right hook point \\
& MaxDist (L5)  & Maximum perpendicular distance between the cow body surface curve and line L5 connecting spike B and right pin point \\
& MaxDist (L6)  & Maximum perpendicular distance between the cow body surface curve and line L6 connecting spike B and left pin point \\
& MaxDist (L7)  & Maximum perpendicular distance between the cow body surface curve and line L7 connecting left hook and left pin point \\
& MaxDist (L8)  & Maximum perpendicular distance between the cow body surface curve and line L8 connecting left hook and spike C point  \\
& MaxDist (L9)  & Maximum perpendicular distance between the cow body surface curve and line L9 connecting left hook and spike B point  \\
& MaxDist (L10) & Maximum perpendicular distance between the cow body surface curve and line L10 connecting left hook and spike A point \\

\hline

\multirow{10}{*}{Area}
& Area (L1)  & Area enclosed between the cow body surface curve and line L1 connecting spike A and the right hook point\\
& Area (L2)  & Area enclosed between the cow body surface curve and line L2 connecting spike B and the right hook point \\
& Area (L3)  & Area enclosed between the cow body surface curve and line L3 connecting spike C and the right hook point \\
& Area (L4)  & Area enclosed between the cow body surface curve and line L4 connecting right pin and the right hook point \\
& Area (L5)  & Area enclosed between the cow body surface curve and line L5 connecting spike B and right pin point \\
& Area (L6)  & Area enclosed between the cow body surface curve and line L6 connecting spike B and left pin point \\
& Area (L7)  & Area enclosed between the cow body surface curve and line L7 connecting left hook and left pin point \\
& Area (L8)  & Area enclosed between the cow body surface curve and line L8 connecting left hook and spike C point \\
& Area (L9)  & Area enclosed between the cow body surface curve and line L9 connecting left hook and spike B point \\
& Area (L10) & Area enclosed between the cow body surface curve and line L10 connecting left hook and spike A point \\

\hline

\multirow{4}{*}{Volume}
& Volume (V1) & Volume enclosed between the cow body surface and the triangular plane defined by spike A, spike B, and the right hook point \\
& Volume (V2) &  Volume enclosed between the cow body surface and the triangular plane defined by spike A, spike B, and the left hook point\\
& Volume (V3) &  Volume enclosed between the cow body surface and the triangular plane defined by spike B, spike C, and the right hook point\\
& Volume (V4) &  Volume enclosed between the cow body surface and the triangular plane defined by spike B, spike C, and the left hook point\\

\hline
\end{tabular}}
\end{table}


\end{document}